\documentclass{article}
\usepackage{times}
\usepackage{graphicx} % more modern
\usepackage{natbib}

\usepackage{amssymb}
\usepackage{amsmath} 
\usepackage{mathrsfs}
\usepackage{mathtools}
\usepackage{bigdelim}
\usepackage{pgfplots}
\usepackage{enumerate}
\usepackage{bm}         %For bolding with greek letters
\usepackage{accents}

\usepackage{rotating}
\usepackage{verbatim}
\usepackage{mathtools}
\usepackage{hhline}
\usepackage{algorithm}
\usepackage{algorithmic}
\usepackage{booktabs}
\usepackage{multirow}
\usepackage{xcolor}
\usepackage{colortbl}

\usepackage{hyperref}

\usepackage[accepted]{icml2017}

\icmltitlerunning{Semi-Markov-Modulated Marked Hawkes Processes for Risk Prognosis}% in Critical Care}

\begin{document} 

\twocolumn[
\icmltitle{Learning from Clinical Judgments: Semi-Markov-Modulated Marked \\ Hawkes Processes for Risk Prognosis}% in Critical Care}



% you can specify symbols, otherwise they are numbered in order
\icmlsetsymbol{equal}{*}

\begin{icmlauthorlist}
\icmlauthor{Ahmed M. Alaa,}{U1}
\icmlauthor{Scott Hu,}{U1}
\icmlauthor{Mihaela van der Schaar}{U1,U2}
\end{icmlauthorlist}

\icmlaffiliation{U1}{University of California, Los Angeles, US}
\icmlaffiliation{U2}{University of Oxford, UK}

\icmlcorrespondingauthor{Ahmed M. Alaa}{ahmedmalaa@ucla.edu}
%\icmlcorrespondingauthor{Eee Pppp}{ep@eden.co.uk}

\icmlkeywords{boring formatting information, machine learning, ICML}

\vskip 0.3in
]



\printAffiliationsAndNotice{\icmlEqualContribution} % otherwise use the standard text.
%\footnotetext{hi}

\begin{abstract} 
Critically ill patients in regular wards are vulnerable to unanticipated adverse events which require prompt transfer to the intensive care unit (ICU). To allow for accurate prognosis of deteriorating patients, we develop a novel continuous-time probabilistic model for a monitored patient's temporal sequence of physiological data. Our model captures ``{\it informatively sampled}" patient {\it episodes}: the clinicians' decisions on when to observe a hospitalized patient's vital signs and lab tests over time are represented by a {\it marked Hawkes process}, with intensity parameters that are modulated by the patient's latent {\it clinical states}, and with observable physiological data (mark process) modeled as a {\it switching multi-task Gaussian process}. In addition, our model captures ``{\it informatively censored}" patient episodes by representing the patient's latent clinical states as an {\it absorbing} semi-Markov jump process. The model parameters are learned from offline patient episodes in the electronic health records via an EM-based algorithm. Experiments conducted on a cohort of patients admitted to a major medical center over a 3-year period show that risk prognosis based on our model significantly outperforms the currently deployed medical risk scores and other baseline machine learning algorithms. 
\end{abstract} 

\section{Introduction} 
\label{intro}
Hospitalized patients are vulnerable to a wide range of adverse events, including cardiopulmonary arrests \cite{kause2004comparison, hogan2012preventable, ForecastICU}, acute respiratory failures \cite{mokart2013delayed}, septic shocks \cite{henry2015targeted}, and post-operative complications \cite{clifton2012gaussian}. For a patient in a regular ward, the occurrence of any such event entails an unplanned transfer to an intensive care unit (ICU), the timing of which is a major determinant of the eventual outcome. Indeed, recent medical studies have confirmed that delayed transfer to the ICU is strongly correlated with morbidity and mortality \cite{Datades1, mokart2013delayed}. The problem of delayed ICU transfer is enormous and acute: over 750,000 septic shocks and 200,000 cardiac arrests occur in the U.S. each year with mortality rates of 28.6$\%$ and 75$\%$ respectively \cite{merchant2011incidence,kumar2011nationwide}. Fortunately, experts believe that much of these events could be prevented with accurate {\it prognosis} and early warning \cite{nguyen2007implementation}. 

Motivated by the proliferation of electronic health records (EHRs) (currently available in more than 75$\%$ of hospitals in the U.S. \cite{charles2016electronic}) we develop a {\it data-driven} real-time risk score that can promptly assess a hospitalized patient's risk of clinical deterioration. Our risk score hinges on a novel continuous-time {\it semi-Markov-modulated marked Hawkes process} model for a monitored patient's {\it episode}, i.e. the patient's evolving (latent) clinical states and her corresponding (observed) physiological data. With the guidance of critical care experts, we conducted experiments on a dataset for a cohort of critically-ill patients admitted to a major academic medical center. Results show that our risk score offers significant gains in the accuracy (and timeliness) of predicting clinical deterioration; our risk score attains a 23$\%$ improvement in the Area Under Receiver Operating Characteristic (AUROC) as compared to the technology currently deployed in our medical center. Since it confers a significant prognostic value in subacute care in wards, the proposed risk score is currently being installed in our medical center as a replacement for the current technology.    

The proposed probabilistic model (based on which our risk score is computed) captures a hospitalized patient's entire {\it episode} as recorded in the EHR. A typical (critical care) patient episode comprises the time of her admission to the ward, the time of her admission to the ICU or discharge from the ward, and a temporal sequence of irregularly sampled physiological data that are collected during her stay in the ward \cite{johnson2016mimic, ghassemi2015multivariate}. We model a patient's episode as being driven by a latent {\it clinical state process}, which we represent as a {\it semi-Markov jump process} \cite{yu2010hidden}, describing the evolution of the patient's ``severity of illness" over time. All the observable physiological variables are modulated by this process: the times at which clinicians decide to observe the patient's physiological data are drawn from a {\it Hawkes point process} \cite{hawkes1974cluster}, the intensity of which is modulated by the patient's clinical state process, whereas the observed physiological data is drawn from a {\it switching multi-task Gaussian process} with hyper-parameters that depend on the patient's clinical state. The patient episode is thus a {\it marked Hawkes process} --with the physiological data serving as the {\it marks}-- that is modulated by the patient's clinical states. We provide a detailed description of the model in Section \ref{Hawk}, and then propose an EM-based algorithm for learning its parameters from the EHR data in Section \ref{Infr}.  

A distinctive feature of our model is its ability to incorporate informative {\it clinical judgments} into the generative process for the patient's episode. The manifestation of informative clinical judgments in the EHR episodes is double-faceted: the patients' episodes are both ``{\it informatively sampled}" and ``{\it informatively censored}". Informative sampling results from the fact that clinicians decide to observe the patient's physiological data more intensely if they believe that the patient is in a ``bad" clinical state \cite{moskovitch2015outcomes, qin2015auxiliary}-- a belief that is based on either the clinician's own assessment of the patient's state, or the communication between the patient and the ward staff \cite{kyriacos2014monitoring}. Informative censoring results from the clinicians' decision on when to send the patient to the ICU or discharge her from the ward, which is indeed informative of the `` clinical deterioration" or ``clinical stability" onsets. In our model, informative censoring is taken into account by adopting an {\it absorbing} semi-Markov chain as a model for the patient's latent clinical states; a patient's risk score at any point of time is thus defined as the {\it probability of eventual absorption} in a ``clinical deterioration" state.     

{\bf Related work:} Marked point processes have been recently used in a very different context to model {\it check-in} data \cite{du2016recurrent, pan2016markov}, but we are not aware of any attempts for their deployment in the medical context. Most of the previous works on risk prognosis for critical care patients viewed informative censoring as a ``{\it surrogate label}" for a patient's clinical deterioration, and hence used those labels to train a supervised (regression) model using the physiological data in a fixed-size time window before censoring. The supervised models used in the literature included logistic regression \cite{ho2012imputation, saria2010integration}, SVMs \cite{wiens2012patient}, Gaussian processes \cite{ghassemi2015multivariate, ForecastICU} and recurrent neural networks \cite{che2016recurrent}. The main limitation of this approach is that, in addition to the fact that it generally does not deal with informatively sampled episodes, it does not model the entire patient's physiological trajectory, and hence it does not accurately capture intermediate (subtle) deterioration stages that are indicative of future severe deterioration stages, which leads to a sluggish risk assessment and delayed ICU alarms. 

Another strand of literature has focused on building probabilistic models, usually variants of Hidden Markov Models (HMMs), for the entire patient's physiological trajectories; applications have ranged from disease progression modeling to neonatal sepsis prediction \cite{wang2014unsupervised, stanculescu2014autoregressive}. These models are not capable of dealing with irregularly-sampled data, do not deal with informatively sampled episodes, and are restricted to the Markovianity assumption which entails unrealistically memoryless clinical state transitions. In \cite{henry2015targeted}, a ranking algorithm was used to construct a risk score for sepsis shocks; however, the approach therein requires the clinicians to provide assessments that order the disease severity at different time instances-- we typically do not have such data in the EHR for ward patients.  

Various works in the medical literature have proposed ``expert-based" medical risk scores for prognosis in hospital wards \cite{morgan1997early, parshuram2009development}, some of which are currently used in practice. The medical literature has also suggested the use of mortality risk scores that are normally used in the ICU, such as APACHE-II and SOFA, as risk scores for ward patients \cite{yu2014comparison}. However, recent systematic reviews have demonstrated the modest net clinical utility of all these scores \cite{mainresult1}. More recently, a data-driven medical risk score based on a simple regression model, known as the {\it Rothman index}, has been developed and commercialized \cite{finlay2014measuring, rothman2013development}. The Rothman index is currently deployed in various major hospitals in the U.S. including our medical center; in Section \ref{Expr}, we show that our risk score significantly outperforms the Rothman index in terms of AUROC and timeliness of ICU admission alarms. 
%; in particular, widely used medical risk scores such as {\it MEWS} \cite{finlay2014measuring} exhibit high false alarm rates which lead to {\it alarm fatigue}, and an unnecessary increase in nursing loads \cite{subbe2001validation}

\section{Structure of the EHR Data} 
\label{EHR}
The subacute care data in an EHR typically comprises a set of {\it episodes}; each episode is a sequence of vital signs and lab tests (physiological data) that have been gathered (by clinicians) for a hospitalized patient at irregularly spaced time instances during her stay in a ward. The episode starts at the time of admission to the ward, and is concluded by either an {\it unplanned admission to the ICU}, which means that the patient was {\it clinically deteriorating}, or a {\it discharge from the ward}, which means that the patient was {\it clinically stable}. We denote an EHR dataset $\mathcal{D}$ that comprises the episodes for $D$ patients as $\mathcal{D} = \{\mathcal{E}^{d}\}_{d=1}^{D},$ where $\mathcal{E}^{d}$ is the episode for the $d^{th}$ patient, and is defined as $\mathcal{E}^{d} = (\{y^d_m, t^d_m\}_{m=1}^{M_d}, T_c^d, l^d),$ with $y^d_m$ being the $m^{th}$ $Q$-dimensional physiological variable (vital signs and lab test outcomes) for patient $d$ observed at time $t^d_m$, and the total number of samples observed for that patient during her episode is $M_d$. The duration of patient $d$'s stay in the ward is denoted by $T_c^d$, whereas her endpoint outcome (clinical deterioration and ICU admission, or clinical stability and discharge) is declared via a binary variable $l^d$; the realization $l^d = 1$ means that patient $d$ was admitted to the ICU, and $l^d=0$ means that the patient was discharged home. We stress that the labels $l^d$ associated with every episode are neither {\it noisy} nor {\it entirely subjective}: we assign a label $l^d = 1$ for patients who actually needed a therapeutic intervention {\it in} the ICU after an {\it unplanned} admission, and assign a label $l^d = 0$ for patients who were discharged and not re-admitted shortly after. We excluded all post-surgical ward patients for whom an ICU admission was preordained since for those patients the prognosis problem is not relevant. Our dataset comprises thousands of episodes for patients admitted to a large medical center over a 3-year period; all the episodes display the structure described above. %Note that while we will conduct experiments on a confidential dataset that belongs to our medical center's EHR (Section \ref{Expr}), our generative model applies to any critical care EHR, including the MIMIC-III public dataset \cite{johnson2016mimic}. 

\begin{figure}[t]
        \centering
        \includegraphics[width=3.5in]{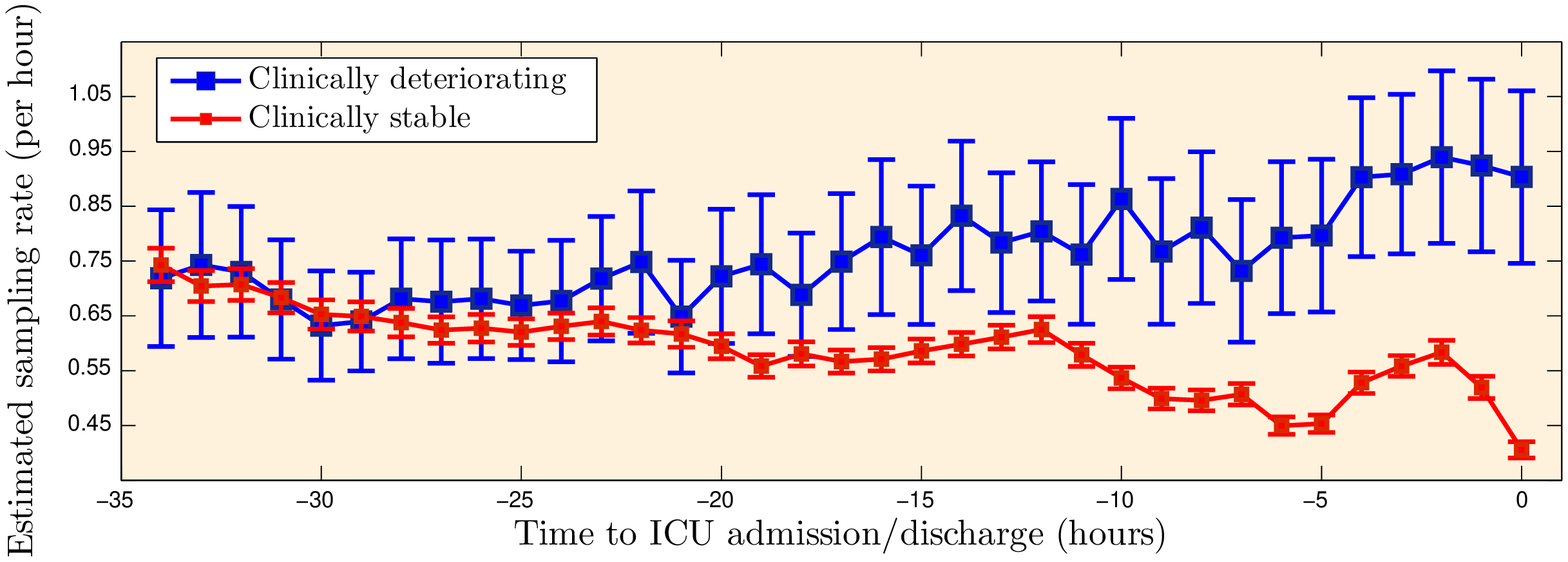}
        \caption{\small Estimated sampling rates (with 95$\%$ confidence intervals) for the physiological data over time.}
\label{Fiq1}
\end{figure}

{\bf The clinicians were right!}\\ % Clarify more that they help as surrogate state estimators
Clinical judgments manifest in the $d^{th}$ episode of $\mathcal{D}$ through informative sampling (encoded in the observation times $\{t^d_m\}_{m=1}^{M_d}$), and informative censoring (encoded in the episode duration $T_c^d$ and the endpoint outcome $l^d$). Figure \ref{Fiq1} is a depiction for both informative sampling and informative censoring. In Figure \ref{Fiq1}, we estimate the physiological data (time-varying) sampling rate using all the episodes in our dataset over a time horizon of 35 hours before ICU admission for deteriorating patients (i.e. patients with $l^d = 1$, with $t=0$ being the ICU admission time), and we compute the same estimates for stable patients (i.e. patients with $l^d = 0$, with $t=0$ being the discharge time). 

We can see from the trends in Figure \ref{Fiq1} that as the deteriorating patient approaches the ICU admission time, the clinicians tend to sample her physiological data more intensely, whereas as the stable patient approaches the discharge time, the clinicians tend to have a more relaxed schedule for observing her vital signs and lab tests. The divergence between the sampling rates for deteriorating and stable patient groups increases as the patients approach their ICU admission and discharge onsets-- that is because the clinicians become less uncertain about the patient's state as time progresses. We have tested the hypothesis that the sampling rate of deteriorating patients is --on average-- larger than that for stable patients in the last 24 hours before ICU admission or discharge via a {\it two-sample $t$-test} with a significance level of 0.05, and the hypothesis was accepted.   

The take-away from Figure \ref{Fiq1} is that the clinician's judgment of the patient's clinical state --manifesting in the vital signs and lab tests sampling rates-- is very predictive of the endpoint outcomes. This implies that there is a room for ``learning from the informative clinical judgments"; that is, one can infer the patients' latent states over time by using the clinicians' observable sampling patterns as proximal noisy labels for those latent states. 
   
\begin{figure*}[t]
        \centering
        \includegraphics[width=6in]{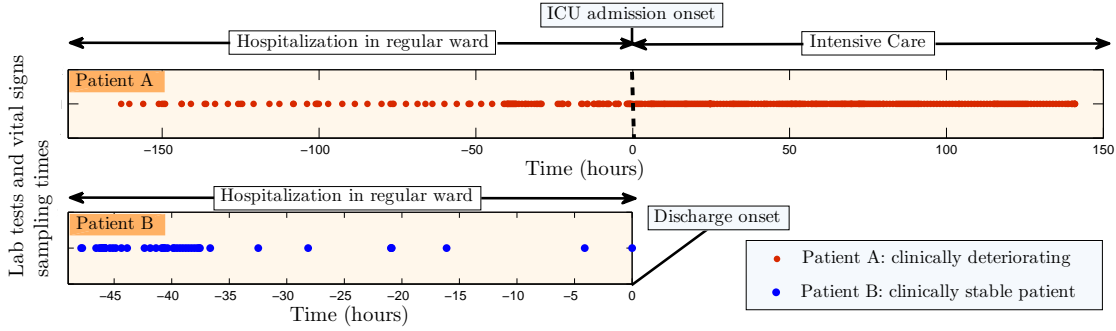}
        \caption{\small Depiction of the physiological data sampling times for a clinically deteriorating and a clinically stable patient.}
\label{Fiq2}
\end{figure*}
   
\section{The Semi-Markov-Modulated Marked Hawkes Process Model}
\label{Hawk}
Now we present a probabilistic model for the episodes $\{\mathcal{E}^d\}_{d=1}^{D}$ that captures the informative sampling and censoring effects discussed in Section \ref{EHR}. We start by modeling the patient's latent clinical state process in Subsection \ref{LCS}, before modeling the observable variables in Subsection \ref{OHWK}.  
\subsection{Latent Clinical States}
\label{LCS}
We assume that each patient's episode is governed by an underlying {\it latent clinical state process} $X(t)$ that represents the evolution of her ``clinical well-being" over time. The patient's latent clinical state $X(t)$ at any point of time $t \in \mathbb{R}_{+}$ ($t=0$ corresponds to the time of admission to the ward) belongs to a finite space $\mathcal{X}$ comprising $N$ states, i.e. $\mathcal{X} = \{1,2,.\,.,.,N\}$. We model informative censoring by assuming that states 1 and $N$ are {\it absorbing states}; state 1 is the state of {\it clinically stability}, at which the patient can be safely discharged home, whereas state $N$ is the state of {\it clinical deterioration}, at which the patient needs to be admitted to the ICU. Whenever the patient is in state $1$ or $N$, her episode is terminated by the clinicians shortly after. All other states in $\mathcal{X}/\{1,N\}$ are {\it transient} states in which the patient needs vigilant monitoring by the ward staff. 

The clinical state process $X(t)$ is a {\it semi-Markov jump process} \cite{yu2010hidden}, i.e. $X(t) = \sum_{n=1}^{K}X_n \, \cdot \, {\bf 1}_{\{\tau_{n} \leq t < \tau_{n+1}\}}$ is a semi-Markov process for which every new state realization $X_n$ starts at a jump time $\tau_n$, where $\tau_1=0$, and lasts for a random sojourn time $S_n = \tau_{n+1}-\tau_n$. A total number of $K$ states are realized in the path $X(t)$, where $K$ is indeed random, and $X_K \in \{1,N\},$ i.e. the patient's episode is concluded by either clinical deterioration or stability. The advantage of adopting a semi-Markovian model for the clinical state process is that unlike Markovian models, semi-Markovianity does not imply memoryless transitions-- the transition probability from one clinical state to another at any time depends on the time spent in the current state, a property that has been recently validated in various clinical state models \cite{taghipour2013parameter}. We adopt an {\it explicit-duration} model for the state sojourn times \cite{johnson2013bayesian}: the $n^{th}$ state sojourn time $S_n$ is drawn from a Gamma distribution\footnote{We model the sojourn time via a Gamma distribution since it encompasses memoryless exponential distributions of Markov models as a special case \cite{liu2015efficient}.} with state-specific parameters as follows
\begin{equation}
S_n|(X_n = i) \sim \mbox{Gamma}(\gamma_{i}), \forall i \in \mathcal{X}. 
\label{eq0}
\end{equation}
Since the model includes two absorbing states ($1$ and $N$) for which the notion of sojourn time is inapplicable, we define the variables $S_n$ of such states as the clinicians' ``{\it response times}" for admitting patients to the ICU or discharging them upon the clinical deterioration/stability onset. The transitions among the clinical states are governed by a semi-Markov transition kernel matrix ${\bf P} = (p_{ij})_{i,j},$ i.e.
\begin{equation}
\mathbb{P}(X_{n+1} = j| X_{n} = i) = p_{ij}, 
\label{eq1}
\end{equation}
where self-transitions are eliminated for all transient states \cite{yu2010hidden,johnson2013bayesian}, i.e. $p_{ii} = 0, \forall i \in \mathcal{X}/\{1,N\},$ and enforced for the two absorbing states $p_{ii} = 1, \forall i \in \{1,N\}$. The initial state distribution is given by $\pi = (\pi_i)_{i=1}^{N},$ where $\pi_i = \mathbb{P}(X(0)=i)$. Every episode $\mathcal{E}^d$ in an EHR dataset $\mathcal{D}$ is associated with a latent clinical state trajectory $\{X^d_n,S^d_n\}^{K^d}_{n=1}$, but we can only observe the absorbing state realization $X_{K^d} = l^d$ in the EHR data (informative censoring).   

\subsection{Observable Physiological Data}
\label{OHWK}
The patient's latent clinical state process $X(t)$ manifests in two ways: (1) it modulates the intensity of sampling the patient's physiological variables (informative sampling), and (2) it modulates the distributional properties of the observed physiological variables. We capture these two effects via a {\it marked point process} model for the patient's episode $\mathcal{E}$: the marked point process $\{(y_m,t_m)\}_{m \in \mathbb{N}_{+}}$ comprises an {\it observation process} $\{t_m\}_{m \in \mathbb{N}_{+}}$, which represents the physiological variables' sampling times, and a {\it mark process} $\{y_m\}_{m \in \mathbb{N}_{+}}$, which represents the realized physiological variables at these sampling times. The distributional specifications of our marked point process are given in the following Subsections.    

\subsubsection{The Observation Process} 
We model the observation process generating the physiological variables' observation times $\{t_m\}_{m \in \mathbb{N}_{+}}$ as a {\it doubly stochastic point process} whose intensity, $\lambda(t)$, is a stochastic process modulated by the latent clinical state process $X(t)$. In particular, the observation process $\{t_m\}_{m \in \mathbb{N}_{+}}$ is modeled as a one-dimensional {\it Hawkes process} with a linear {\it self-exciting intensity function} $\lambda(t,X(t))$ \cite{lee2016hawkes}, i.e.
\begin{equation}
\lambda(t,X(t)=i) = \lambda^{o}_{i} + \alpha_{i}\,\sum_{\tau < t_m < t}e^{-\beta_{i} (t-t_m)},  
\label{eq2}
\end{equation}
$\forall\, i \in \mathcal{X},$, where $\lambda^{o}_{i}$, $\alpha_i$ and $\beta_i$ are the state-dependent intensity parameters, $e^{-\beta_{i} (t-t_m)}$ is an exponential {\it triggering kernel}, and $\tau < t$ is the time of the most recent jump in $X(t)$. In order to ensure the {\it local stationarity} of the Hawkes process within the sojourn time of every latent state, we assume that $\frac{\alpha_i}{\beta_i} < 1, \forall i \in \mathcal{X}$ \cite{roueff2016locally}; the expected value of the intensity function is therefore given by $\mathbb{E}[\lambda(t,X(t)=i)] = \frac{\lambda^{o}_{i}}{1-\frac{\alpha_{i}}{\beta_{i}}}$. For $\beta_{i} = \infty$ or $\alpha_{i} = 0$, we recover a modulated {\it Poisson} process as a special case \cite{pan2016markov}.      

In Figure \ref{Fiq2}, we depict the observation times in two patients' episodes: {\it patient A} being a clinically deteriorating patient, and {\it patient B} being a clinically stable patient. We can see that patient A's episode had its sampling rate escalating as her condition was worsening; the sampling rate remained intense after she was admitted to the ICU. On the other hand, patient B's episode exhibited a decelerating sampling rate on her path to clinical stability. We can also notice that the observation times display a subtle ``clustered" pattern that point out to their temporal dependencies-- indeed, the clinicians are not memoryless, and the times at which they observe the physiological data are dependent. In the light of the above, the modulated Hawkes process described above appears to be a sensible model for the observation process $\{t_m\}_{m \in \mathbb{N}_{+}}$ as it captures both the time-varying intensity and the temporal dependencies illustrated in Figure \ref{Fiq2}. 
     
\subsubsection{The Mark Process}
Now we provide the distributional specification of the {\it mark process} $\{y_m\}_{m \in \mathbb{N}_{+}}$. Since the physiological data are irregularly sampled from an underlying continuous-time physiological process $Y(t)$ at the sampling times determined by the {\it observation process} $\{t_m\}_{m \in \mathbb{N}_{+}}$, a convenient model for $Y(t)$ is a {\it switching Gaussian Process} defined as follows: $Y(t) = \sum_{n=1}^{K}Y_n(t) {\bf 1}_{\{\tau_n \leq t < \tau_{n+1}\}},$ with
\begin{equation}
Y_n(t)|(X_n = i) \sim \mathcal{GP}(m_i(t), k_{i}(t,t^{\prime})),
\label{eq3}
\end{equation}  
where $m_i(t)$ and $k_{i}(t,t^{\prime})$ are the state-dependent mean function and covariance kernel, respectively. We use a constant mean function $m_i(t) = m_i$ and a {\it Mat\'{e}rn kernel} given by
\begin{equation}
k_{i}(t,t^{\prime}) = \frac{\left(\frac{\sqrt{2v_i-1}\,|t-t^{\prime}|}{\ell_i}\right)^{v_i-\frac{1}{2}}\,K_{v_i-\frac{1}{2}}\left(\frac{\sqrt{2v_i-1}\,|t-t^{\prime}|}{\ell_i}\right)}{2^{v_i-\frac{3}{2}}\,\Gamma(v_i-\frac{1}{2})},
\label{eq3}
\end{equation}
where $v_i \in \mathbb{N}_{+}$, $\ell_i \in \mathbb{R}_{+}$, $\Gamma(.)$ is the {\it Gamma function} and $K_{v_i-\frac{1}{2}}(.)$ is a {\it modified Bessel function} \cite{rasmussen2006gaussian}.

Our choice for the {\it Mat\'{e}rn kernel} is motivated by its ability to represent various commonly used stochastic processes; for instance, when $v_i = 1$, then $Y_n(t)|(X_n = i)$ is an {\it Ornstein-Uhlenbeck} process \cite{rasmussen2006gaussian}, whereas for a general integer value of $v_i$, $Y_n(t)|(X_n = i)$ is a continuous-time analogue of the {\it Auto-regressive process} $\mbox{AR}(v_i)$-- a process that has been widely used to model physiological time-series data \cite{stanculescu2014autoregressive}. By constructing $Y_n(t)$ as a continuous-time analog of the AR model, the process $Y(t) = \sum_{n=1}^{K}Y_n(t) {\bf 1}_{\{\tau_n \leq t \leq \tau_{n+1}\}}$ becomes a continuous-time {\it switching AR model} that is modulated by the patient's latent clinical state process $X(t)$. We observe the continuous-time process $Y(t)$ only at the sampling times dictated by the observation process $(t_m)_{m \in \mathbb{N}_{+}}$, and the resulting process $\{y_m\}_{m \in \mathbb{N}_{+}}$ defines the mark process. The observation process together with the mark process, both modulated by the latent clinical state process $X(t)$, constitute a {\it marked Hawkes process}, which completely describes a patient's episode.          
 
\begin{figure}[t]
        \centering
        \includegraphics[width=3.5in]{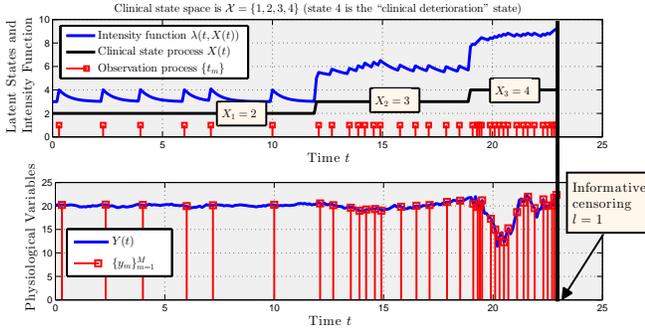}
        \caption{\small An episode sampled from the proposed model.}
\label{Fiq3}
\end{figure}

The mark process defined above is one-dimensional, and hence we need to extend the definition to handle a multi-dimensional process that represents multiple lab tests and vital signs. In other words, we seek a continuous-time analog of the {\it switching Vector Auto-regressive (VAR) model} rather than an AR model\footnote{In Appendix A of the supplementary material, we establish the connection between the switching multi-task Gaussian Process model described herein and the conventional VAR model, showing that the former is the continuous-time analog of the latter.}. This is achieved by adopting a {\it multi-task Gaussian Process} as a model for the $Q$-dimensional physiological process $Y(t) \in \mathbb{R}^{Q}$ \cite{durichen2015multitask}. That is, we assume that $Y_n(t)|(X_n = i) \sim \mathcal{GP}(m_i(t), {\bf K}_{i}(t,t^{\prime})),$ where the covariance kernel ${\bf K}_{i}(t,t^{\prime}) = \{k_i(r,g,t,t^{\prime})\}^{Q}_{r,g = 1}$ is based on the {\it intrinsic correlation model} \cite{bonilla2007multi}, i.e. ${\bf K}_{i}(t,t^{\prime})$ can be written in the following separable form
\begin{equation}
k_i(r,g,t,t^{\prime}) = {\bf \Sigma}_{i}(r,g)\,\cdot\,k_i(t,t^{\prime}),
\label{eq4}
\end{equation}
where $k_i(r,g,t,t^{\prime})$ is the covariance between the $r^{th}$ physiological variable at time $t$ and the $g^{th}$ physiological variable at time $t^{\prime}$, ${\bf \Sigma}_{i}$ is a $Q \times Q$ {\it intrinsic correlation matrix}, and $k_i(t,t^{\prime})$ is the Mat\'{e}rn kernel in (\ref{eq3}). For every state $i$, we denote the multi-task GP parameter set as $\Theta_i = (m_i(t), {\bf K}_{i}(t,t^{\prime}))$, and the Hawkes process parameter set as $\Lambda_i$. The entire model parameters can be bundled in the parameter set $\Omega$ as follows
\[\Omega = \{{\bf P},(\pi_{i}, \gamma_i, \Lambda_i, \Theta_i)_{i \in \mathcal{X}}\}.\]
Given a parameter set $\Omega$, we can easily generate sample patient episodes from our model by first sampling a state sequence $\{X_1,.\,.\,.,X_K\}$ using the semi-Markov transition kernel, then sampling a corresponding sequence of sojourn times $\{S_1,.\,.\,.,S_K\}$ from the state-dependent Gamma distributions, then sampling a set of multi-task Gaussian process $\{Y_1(t),.\,.\,.,Y_K(t)\}$, and finally sampling a sequence of observation times $\{t_m\}_{m=1}^{M}$ using Ogata's {\it modified thinning algorithm} \cite{ogata1981lewis}. Figure \ref{Fiq3} depicts one patient episode sampled from our model. (An algorithm for sampling episodes from our model is provided in Appendix B in the supplementary material.)

\section{Learning and Inference}
\label{Infr}
In this Section, we develop an {\it offline} learning algorithm that learns the model parameter $\Omega$ using the offline training episodes in $\mathcal{D}$, and a {\it real-time} risk scoring algorithm that computes a hospitalized patient's risk over time. The learning algorithm operates by first detecting change-points in the physiological data and the observation process; using the detected change-points, the algorithm segments each episode into a sequence of states, and uses an EM algorithm to estimate the model parameters. The real-time risk scoring algorithm operates by inferring the patient's current state via forward-filtering, and then computing the probability of eventual absorption in the deteriorating state.  
  
\subsection{The Offline Learning Algorithm}
\label{Offlearn}
Estimating the model parameters $\Omega$ from the dataset $\mathcal{D} = \{\mathcal{E}^d\}_{d=1}^D$ is a daunting task due to the hiddenness of the patients' clinical state trajectories $\{X^d_n,S^d_n\}_{d=1}^D$; an application of MCMC-based inference methods, such as the method in \cite{qin2015auxiliary}, will incur an excessive computational cost for a complex model like ours. We therefore developed an efficient three-step learning algorithm that capitalizes on the structure of the patients' episodes in order to find a point estimate $\hat{\Omega}$ for $\Omega$. The three steps of the our learning algorithm are listed hereunder\footnote{The details for all the algorithms in this Section are provided in Appendix C in the supplementary material.}.      

\textit{\textbf{Step 1: Change-point detection}}\\
We first estimate the jump times $\{\tau^d_1,.\,.\,.,\tau^d_{K^d}\}$ for every episode $\mathcal{E}^d$ in $\mathcal{D}$. This is achieved by using the {\it E-divisive} change-point detection algorithm \cite{matteson2014nonparametric}. Since the E-divisive algorithm is nonparametric, we are able to estimate the onsets of all clinical states, i.e. the jump times of $X^d(t)$, prior to finding the estimate $\hat{\Omega}$. We let the E-divisive algorithm jointly detect changes in the distributions of the observable variables $\{y^d_m\}^{M_d}_{m=1}$ and the observation process $\{t^d_m\}^{M_d}_{m=1}$ by creating an {\it augmented} vector of observables that comprises both the physiological observations and a ``differential" observation process, i.e. 
\begin{equation}
\{\hat{\tau}^d_1,.\,.\,.,\hat{\tau}^d_{K^d}\} = \mbox{E-divisive}((y^d_1, \Delta t^d_1),.\,.\,.,(y^d_{M^d}, \Delta t^d_{M^d})), \nonumber
\end{equation}
where $\Delta t^d_m = t^d_m - t^d_{m-1}$, with $\Delta t^d_1 = 0$. By the end of this step, we obtain an estimate for the start and end times of all clinical state realizations for every episode $\mathcal{E}^d$.   

\textit{\textbf{Step 2: Maximum Likelihood Estimation (MLE) of the absorbing states' parameters}}\\
By virtue of informative censoring, we know the identities of all the absorbing states, i.e. $X^d_{K^d} = l^d, \forall d$. Now that we have estimates for the onsets of the absorbing states, obtained from step 1, then we can estimate the response times as $\hat{S}^d_{K^d} = T_c^d-\hat{\tau}^d_{K^d}, \forall 1 \leq d \leq D$. Define the (fully observable) sub-dataset $\mathcal{D}_{i}$ as follows
\begin{equation}
\mathcal{D}_{i} = \left\{(y_m, t_m)_{\{t_m \geq \hat{\tau}^d_{K^d}\}}, \hat{S}^d_{K^d}: l^d = i, \mathcal{E}^d \in \mathcal{D}\right\}, \nonumber 
\end{equation}
$\forall i \in \{0,1\}$. Given such a fully fledged specification of the absorbing states' onsets, identities, and the corresponding physiological variables, we can directly apply MLE to estimate the parameters $\Theta_1, \Theta_N, \gamma_1, \gamma_N, \Lambda_1$ and $\Lambda_N$. Using the dataset $\mathcal{D}_{i}$, the parameter $\Theta_i$ is estimated using the gradient method for Gaussian processes as in \cite{bonilla2007multi}, $\gamma_i$ is estimated using the standard MLE estimating equations, and $\Lambda_i$ is estimated by maximizing the recursive likelihood formula in \cite{ogata1981lewis} using the Nelder-Mead simplex method \cite{nelder1965simplex}. 

\textit{\textbf{Step 3: Estimation of the transient states' parameters using the EM algorithm}}\\
While the absorbing states are observable, the transient states are all hidden. In order to estimate the parameters ${\bf P}, \{\Theta_i\}_{i=2}^{N-1}, \{\gamma_i\}_{i=2}^{N-1}$, and $\{\Lambda_i\}_{i=2}^{N-1}$, we use the jump times' estimates $\{\hat{\tau}^d_1,.\,.\,.,\hat{\tau}^d_{K^d-1}\}$ (obtained from step 1) in order to {\it segment} every episode $d$ into a set of finite transition, and hence we obtain a discrete-time HMM-like process. We truncate all the episodes by removing the data belonging to the absorbing state, and run the EM algorithm in order to estimate the transient state parameters. The EM algorithm takes advantage of informative censoring in the forward-backward message passing stage by computing the backward messages conditioned on the identity of the endpoint absorbing state $l^d$ for every episode $d$. 

\subsection{The Real-time Risk Scoring Algorithm}
\label{rscore} 
Having learned the model parameters $\hat{\Omega}$ from an offline dataset $\mathcal{D}$, we now explain how risk scoring is conducted in real-time for a newly hospitalized patient. The patient risk score at time $t$ is denoted by $R(t)$, and is defined as $R(t) = \mathbb{P}(X(\infty) = N\,|\,\{y_m,t_m\}, t_m \leq t, \hat{\Omega})$. That is, the risk score $R(t)$ is the probability of being eventually absorbed in the deteriorating state $N$ given the observable physiological data up to time $t$. Using Bayes' rule, the risk score $R(t)$ is given by  
\[\sum_{i \in \mathcal{X}} \underbrace{\mathbb{P}(X(t) = i\,|\,\{y_m,t_m\})}_{\small \mbox{Current state}}\,\cdot\, \underbrace{\mathbb{P}(X(\infty) = N\,|\,X(t)=i)}_{\small \mbox{Future transition}},\]
where the ``current state" term is computed in real-time using the efficient (dynamic programming) {\it forward-filtering} algorithm, whereas the ``future transition" term is computed offline using the estimated model parameters. 

\section{Experiments}
\label{Expr}
In order to evaluate the prognostic utility of our model, we conducted experiments on a dataset $\mathcal{D}$ comprising information on patient admissions to a major medical center over a 3-year period, and compared the proposed risk score defined in Subsection \ref{rscore} with other competing baselines. We briefly describe our dataset in the next Subsection, and then present the experimental results. A very detailed description for our dataset and the implementation of the baselines is provided in Appendix D in the supplementary material.  
\subsection{Data Description}
 Each patient record in $\mathcal{D}$ is an episode that is formatted as described in Section \ref{EHR}. The patients' cohort in $\mathcal{D}$ is very heterogeneous; diagnoses included sepsis, hypertension, renal failure, leukemia, septicemia and pneumonia. Each patient's episode in $\mathcal{D}$ comprises 21 vital signs and lab tests that are collected for the patient over time, along with the time instances at which they where collected. The vital signs include diastolic and systolic blood pressure, Glasgow coma scale score, heart rate, eye opening, respiratory rate, temperature, $O_2$ saturation and device assistance, best motor and verbal responses. The lab tests included measurements of chloride, glucose, urea nitrogen, white blood cell count, creatinine, hemoglobin, platelet count, potassium, sodium and $CO_2$. In all the experiments conducted in this Section, we split $\mathcal{D}$ into a training set consisting of admissions over a 2.5-year period (5,000 episodes) and a testing set consisting of admissions over a 6-month period (1,094 episodes).

\subsection{Results}
We ran the offline learning algorithm in Subsection \ref{Offlearn} (with 1000 EM iterations) on the 5,000 training episodes in $\mathcal{D}$, and obtained an estimate $\hat{\Omega}$ for the semi-Markov-modulated marked Hawkes process that describes the patient cohort. Using the {\it Bayesian information criterion}, we selected an instantiation of our model with 4 clinical states, where state 1 is the {\it clinical stability} (absorbing) state, and state 4 is the {\it clinical deterioration} (absorbing) state. The learned Hawkes process intensity functions for these two states are given by 
\begin{align}
\lambda(t,1) &= \underbrace{0.55}_{\mbox{\small Baseline intensity}\,\, \downarrow} + \underbrace{0.2\,\sum_{t_m}e^{-8.46 (t-t_m)}}_{\mbox{\small Temporal dependencies}\,\, \downarrow} , \nonumber \\
\lambda(t,4) &= \underbrace{0.82}_{\mbox{\small Baseline intensity}\,\, \uparrow} + \underbrace{0.16\,\sum_{t_m}e^{-1.36 (t-t_m)}}_{\mbox{\small Temporal dependencies}\,\, \uparrow} , \nonumber
\end{align}
where $\lambda(t,X(t))$ is measured in {\it samples per hour}. We note that the estimated Hawkes process parameters accurately describe the clinicians' judgments; when the patient is in the deteriorating state (state 4), the clinicians tend to observe her physiological measurements more frequently. This manifests in the baseline intensity of $\lambda(t,4)$ being $50\%$ higher than that of $\lambda(t,1)$. Moreover, we note that when the patient is clinically stable (state 1), the temporal dependencies between the observation times almost disappear as the exponential triggering kernel plays little role in determining the observation times, i.e. $\lambda(t,1) \approx 0.55$, which renders the observation process closer to a Poisson process. Contrarily, when the patient is deteriorating, strong temporal dependencies are displayed in the observation times-- this is intuitive since for a deteriorating patient, the follow-up times decided by the clinicians strongly depend on what have been observed in the past. These distinguishing state-specific features of the clinical judgments are the essence of informative sampling, which allows us to integrate physiological data together with clinical experience while learning the patient's physiological model.   

\begin{figure}[t]
        \centering
        \includegraphics[width=3.5in]{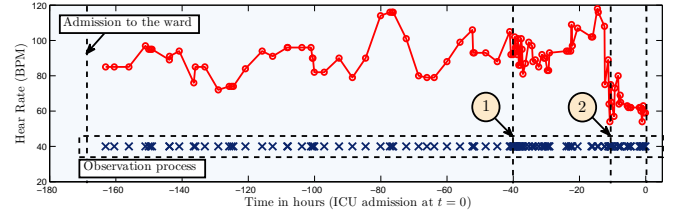}
        \caption{\small An episode for a hospitalized cardiac patient.}
\label{Fiq4}
\end{figure}
We illustrate the value of informative sampling in Figure \ref{Fiq4} through an episode for a cardiac patient who was admitted in the ward for 1 week before being sent to the ICU upon a cardiac arrest. When running the offline learning algorithm in Section \ref{Offlearn} while assuming that the observation process $\{t^d_m\}_m$ is uninformative (i.e. $\lambda(t,i) = \lambda(t,j), \forall i,j \in \mathcal{X}$), the detected clinical deterioration onset is only 10 hours ahead of the cardiac arrest event (onset (2) in Figure \ref{Fiq4}): this is because the states are estimated solely based on the physiological data, and hence the clinical deterioration state is detected only when the patient's heart rate fell below 60 beats per minute (BPM). When running the learning again but with informative sampling taken into account, the detected clinical deterioration onset is 40 hours ahead of the cardiac arrest event (onset (1) in Figure \ref{Fiq4}); this is the time instance at which the clinicians decided to monitor the patient more vigilantly (i.e. more intense sampling rate) even though the evidence for an upcoming cardiac arrest in her heart rate trajectory was rather subtle. The clinicians' decision to intensely observe the patient's physiological trajectory is based either on their experience, or on apparent symptoms or complains from the patient that were not recorded in the EHR. By integrating the clinicians' judgments into our model, we are able to capture the subtleties in the patients' temporal physiological parameters, and hence learn more accurate representations for the clinical states.  
 
We then computed the real-time risk score (as described in Subsection \ref{rscore}) for the testing episodes, and compared its {\it sensitivity-precision} AUCROC with that of the baseline risk scoring methods listed hereunder.\\
\\
\textit{\textbf{Medical Risk Scores:}} we considered the two most commonly used medical risk scores in regular wards-- the MEWS score \cite{morgan1997early} and the Rothman index\footnote{At the time of conducting these experiments, the Rothman index was deployed in more than 60 major hospitals in the US.} \cite{rothman2013development}. We implemented the MEWS score and the Rothman index as described in \cite{kyriacos2014monitoring} and \cite{rothman2013development}, respectively. We also included the APACHE-II and SOFA scores in our comparisons.\\ 
\\ 
\textit{\textbf{Machine Learning Algorithms:}} we considered the traditional approach for real-time risk scoring, which treats informative censoring as a surrogate label based on which a supervised regression model is learned offline, and then risk scoring is applied in real-time using the temporal data within a sliding window-- we call these methods ``sliding-window methods". We implemented sliding-window methods based on logistic regression \cite{ho2012imputation, saria2010integration}, random forests, and Gaussian process regression \cite{ghassemi2015multivariate,ForecastICU}. In addition, we compared our risk score with a standard Hidden Markov Model with Gaussian emissions. The relevant physiological measurements for every baseline were selected through the {\it correlated feature selection} method \cite{yu2003feature}. The hyper-parameters of all the baselines, including the size of the sliding window for the supervised learning methods, were optimized via cross-validation. To handle the irregularly sampled data, we discretized the time horizon into 1-hour steps and fed the baselines with interpolated, discrete-time episodes.  

\begin{table}
    \centering
\begin{tabular}{ |c|c|c| }
\toprule[1.25pt]
\hline
\multicolumn{2}{|c|}{{\bf Method}} & {\bf AUROC} \\ \hline  
\toprule[1.25pt]
\multicolumn{2}{|c|}{{\bf Proposed Score}} & 0.481 \\ \hline
\multirow{4}{*}{{\bf Medical Scores}} & Rothman Index & 0.255 \\ 
 & MEWS & 0.182 \\
 & APACHE-II & 0.130 \\ 
 & SOFA & 0.127 \\ 
\hline
\multirow{4}{*}{{\bf ML Algorithms}} & SW-RF & 0.366 \\
& HMMs & 0.321 \\ 
 & SW-GP & 0.305 \\
& SW-LR & 0.267 \\ 
\hline
\bottomrule[1.25pt]
\end{tabular}
\caption{\small Comparison between various risk scoring methods ($p < 0.01$). (SW: Sliding-window, GP: Gaussian process regression, LR: Logistic Regression, RF: Random Forests)}
\label{Table3ICU}
\end{table}
In Table \ref{Table3ICU}, we compare the performance of our risk score with the baselines in terms of the sensitivity-precision AUROC. As we can see, the proposed risk score offers a 23$\%$ AUROC improvement as compared to the best performing medical risk score --the Rothman index-- which was the score deployed in our medical center at the time of conducting this experiment. Moreover, our risk score also provides significant gains over discriminative sliding-window regression models; the proposed risk score achieves a 11.5$\%$ AUROC improvement as compared to the best performing ML algorithm (random forest). In addition, the proposed risk score achieves a 16$\%$ AUROC improvement as compared to a standard HMM with Gaussian emission variables\footnote{The adoption of a semi-Markovian model for the clinical state process protects our model from the overtly rapid state switching behavior that is introduced by memoryless HMMs \cite{matteson2014nonparametric}.}. On average, our risk score prompts ICU alarms 8 hours before the censoring time at a sensitivity of 50$\%$ and precision of 35$\%$. 

\section{Discussion: Chicken-and-egg}
\label{Disc}
We stress that while computing our risk score for the testing episodes, we did not use the information conveyed in the observation process $\{t^d_m\}_m$ to infer the patients' clinical states. This is because in practice, the value of the real-time risk score $R(t)$ itself influences the clinician's behavior and hence impacts the observation process, creating a {\it chicken-and-egg} dilemma in which one cannot clearly conceptualize the causal relation between the risk score and the clinicians' judgments. A very interesting research direction is to consider an observation process that is modulated by both the patient's state and the real-time risk score through an intensity function $\lambda(t,X(t),R(t))$, where the algorithm learns the clinical state representation {\it online} by ``sharing experience" with the clinicians. That is, the algorithm uses the clinician's judgments to refine its clinical state model, which leads to a refined risk score $R(t)$ that would in turn allow the clinician to exhibit more accurate judgments; an online learning process that would ideally converge to a state of ``shared knowledge" between the clinician and the system. 

The significant prognostic value offered by our risk score promises a great improvement in the quality of subacute care in wards. By utilizing the proposed score instead of the current technology, clinicians in a crowded ward can better focus their attention on patients at real risk of deterioration, and can also plan for timely ICU admissions and effective therapeutic interventions. With the high in-hospital mortality rates in wards, deploying our risk score may help save thousands of lives annually-- we are currently working towards installing the proposed risk score in our medical center.

\bibliography{jmlr_ref4}
\bibliographystyle{icml2017}
\newcounter{defcounter}
\setcounter{defcounter}{0}
\newenvironment{myequation}{%
\addtocounter{equation}{-1}
\refstepcounter{defcounter}
\renewcommand\theequation{A\thedefcounter}
\begin{equation}}
{\end{equation}}

\section*{Supplementary Material}
\subsection*{Appendix A: Multitask Gaussian Processes as Vector Autoregressive Models}
In order to show the equivalence between a multitask GP with a Mat\'{e}rn kernel and the continuous-time VAR process, we first start by writing down the classical discrete-time VAR process ${\bf y}(t)$ in its difference equation form as follows 
\begin{myequation}
{\bf y}(t) = {\bf e}(t) + {\bf A}_1\,\cdot\,{\bf y}(t-1)+\,.\,.\,.+{\bf A}_p\,\cdot\,{\bf y}(t-p),
\label{eqA1}
\end{myequation}
where ${\bf e}(t) \in \mathbb{R}^{Q\times 1}$ is a white noise vector and ${\bf A}_m = [a_{m,ij}]_{ij} \in \mathbb{R}^{Q \times Q}$ is the matrix associated with the $m^{th}$ lag of the process ${\bf y}(t)$. Here we assume, as usual, that the noise process ${\bf e}(t)$ is zero mean with a Dirac-delta auto-correlation function, i.e. $\mathbb{E}[{\bf e}(t)]=0$ and $\mathbb{E}[{\bf e}(t)\,\cdot\,{\bf e}(t-l)]=0,\, \forall \, l \in \mathbb{N}_{+}$. Note that when ${\bf e}(t)$ is assumed to be Gaussian, then ${\bf y}(t)$ is an order-$p$ Gauss-Markov process. The continuous-time version of the VAR process ${\bf y}(t)$ is one that is defined using a {\it stochastic differential equation} (SDE) that is equivalent to the difference equation in (\ref{eqA1}), i.e. 
\begin{myequation}
{\bf y}(t) = {\bf e}(t) + {\bf A}_1\,\cdot\,\frac{\partial {\bf y}(t)}{\partial t}+\,.\,.\,.+{\bf A}_p\,\cdot\,\frac{\partial^p {\bf y}(t)}{\partial t^p},
\label{eqA2}
\end{myequation}
where $\frac{\partial {\bf y}(t)}{\partial t}$ is the component-wise derivative of the vector-valued process ${\bf y}(t)$ with respect to $t$. Now we want to find the {\it power spectrum} $S_y(\omega)$ for the (stationary) process corresponding to the SDE in (\ref{eqA2}). This is achieved by conceptualizing (\ref{eqA2}) as an LTI system with a white noise process as the input and the VAR process ${\bf y}(t)$ as the output, and thus computing the power spectrum as $S_y(\omega) = |H(j\omega)|^2$, where $H(j\omega)$ is the LTI system's transfer function assuming that the noise process ${\bf e}(t)$ has a normalized (unity) power spectrum. In order to proceed with our analysis, we start by taking the Fourier transform for (\ref{eqA2}) as follows  
\begin{myequation}
{\bf Y}(j\omega) = {\bf E}(j\omega) + \sum_{n=1}^{p}(j\omega)^n\,{\bf A}_n\,{\bf Y}(j\omega),
\label{eqA3}
\end{myequation}
which can be further reduced as follows
\begin{myequation}
{\bf Y}(j\omega) = ({\bf I}-\sum_{n=1}^{p}(j\omega)^n\,{\bf A}_n)^{-1}{\bf E}(j\omega).
\label{eqA4}
\end{myequation} 
Based on (\ref{eqA4}), we can easily compute the (matrix-valued) transfer function $H(j\omega) = [h_{mn}(j\omega)]_{mn}$ as follows
\begin{myequation}
H(j\omega)= ({\bf I}-\sum_{n=1}^{p}(j\omega)^n\,{\bf A}_n)^{-1}.
\label{eqA5}
\end{myequation} 
Since we assumed (without loss of generality) that ${\bf E}(j\omega) = {\bf 1}$, then the power spectrum of any component of ${\bf y}(t)$ can thus be found by examining the structure of $S^m_y(\omega) = |\sum_{n} h_{mn}(j\omega)|^2 = (\sum_{n} h_{mn}(j\omega))\,\cdot\,(\sum_{n} h_{mn}(-j\omega))$.

Let us take the case when ${\bf A}_n = \mbox{diag}(a_{n,11},.\,.\,.,a_{n,QQ})$ is a diagonal matrix for every $n$. In this case, we have that
\begin{myequation}
h_{mn}(j\omega) = (\sum_{n=0}^{p}(j\omega)^n\,a_{n,mm})^{-1}, 
\label{eqA6}
\end{myequation} 
and hence we have that
\begin{myequation}
S^m_{y}(j\omega) = \frac{1}{(\sum_{n=0}^{p}(j\omega)^n\,a_{n,mm})\,\cdot\,(\sum_{n=0}^{p}(-j\omega)^n\,a_{n,mm})}. 
\nonumber
\end{myequation} 
From Section (B.2.1) in \cite{rasmussen2006gaussian}, we know that for a particular selection of the coefficients $[a_{n,mm}]$, the power spectrum can be put in the form
\begin{myequation}
S^m_{y}(j\omega) = \frac{1}{(4\pi^2\,\omega^2+\alpha^2)^p}. 
\label{eqA7}
\end{myequation}
By taking the inverse Fourier transform of (\ref{eqA7}), one can see that the corresponding covariance function is of the form $\sum_{n=0}^p \beta_n\,|t|^n\,e^{-\alpha\,|t|}$, which is a special case of the Mat\'{e}rn kernel in (\ref{eq3}). For a general setting of ${\bf A_n}$, one can easily reach to the form $\sum_{n=0}^p \beta_n\,|t|^n\,e^{-\alpha\,|t|}$ as an approximate form for the covariance function by using the fact that $({\bf I}-\sum_{n=1}^{p}(j\omega)^n\,{\bf A}_n)^{-1} = {\bf I} - \frac{1}{1+\mbox{trace}((\sum_{n=1}^{p}(j\omega)^n\,{\bf A}_n)^{-1})}\, (\sum_{n=1}^{p}(j\omega)^n\,{\bf A}_n).$ 

\subsection*{Appendix B: Sampling Episodes from a Semi-Markov Modulated Marked Hawkes Process}
In order to generate a sample for a patient's episode given a parameter set $\Omega$, we first sample a state sequence $\{X_1,.\,.\,.,X_K\}$ using the semi-Markov transition kernel, then sample a corresponding sequence of sojourn times $\{S_1,.\,.\,.,S_K\}$ from the state-dependent Gamma distributions, then sample a set of multi-task Gaussian process $\{Y_1(t),.\,.\,.,Y_K(t)\}$, and finally sampling a sequence of observation times $\{t_m\}_{m=1}^{M}$ using Ogata's {\it modified thinning algorithm} \cite{ogata1981lewis}. Algorithm \ref{alg1} generates samples from a semi-Markov modulated Hawkes Process, whereas algorithm \ref{alg2} is a thinning algorithm used as a sub-module in algorithm \ref{alg1} in order to sample a Hawkes process by thinning an inhomogeneous Poisson process. 

\begin{algorithm}[t]
   \caption{Sampling an episode from a semi-Markov modulated Hawkes Process}
   \label{alg1}
\begin{algorithmic}
   \STATE {\bfseries Input:} parameter set $\Omega$
	 \STATE {\bfseries Output:} an episode $\mathcal{E} = \{(y_m, t_m)_{m=1}^{M}, (X_n, S_n)_{n=1}^{K}\}$
	 \STATE $X_1 = j \sim \pi_j, \tau_1 $
	 \STATE $S_1\,|\,(X_1=j) \sim \mbox{Gamma}(\gamma_{j})$
	 \STATE $Y_1(t)\,|\,(X_1=j) \sim \mathcal{GP}(\Theta_{j})$
	 \STATE $\mathcal{T}_1 \gets \mbox{Thin}(\Lambda_{j},\tau_n,\tau_n+S_n)$
	 \STATE $n \gets 2$
   \REPEAT
   \STATE $X_{n+1} = j|X_n = i \sim p_{ij}$.
	 \STATE $S_n\,|\,(X_n=j) \sim \mbox{Gamma}(\gamma_{j}),\, \tau_n = \tau_{n-1}+S_n$
	 \STATE $Y_n(t)\,|\,(X_n=j) \sim \mathcal{GP}(\Theta_{j})$
	 \STATE $\mathcal{T}_n \gets \mbox{Thin}(\Lambda_{j},\tau_n,\tau_n+S_n)$
	 \STATE $n \gets n+1$
   \UNTIL{$X_n \in \{1,N\}$}
	 \STATE $Y(t) = \sum_{n}Y_n(t)\,\cdot\,{\bf 1}_{\{\tau_{n}\leq t \leq \tau_{n+1}\}}$
	 \STATE $\{t_m\}_{m=1}^{M} \gets \bigcup_n \mathcal{T}_n,\, \{y_m\}_{m=1}^{M} \gets \{Y(t_m)\}_{m=1}^{M}$
\end{algorithmic}
\end{algorithm}

\begin{algorithm}[t]
   \caption{Sampling a Hawkes process by Ortega's modified thinning algorithm $\mbox{Thin}(\Lambda,\tau_o,\tau_1)$}
   \label{alg2}
\begin{algorithmic}
   \STATE {\bfseries Input:} Hawkes process parameters $\Lambda$, interval $[\tau_o,\tau_1]$
	 \STATE {\bfseries Output:} a point process $\{t_m\}$ on the interval $[\tau_o,\tau_1]$
   \STATE $\mathcal{T} = \emptyset, s = 0, n = 0, T = \tau_1-\tau_o$
   \WHILE{$s < T$} 
	 \STATE $\bar{\lambda} = \lambda(S^{+}) = \lambda_o + \alpha\,\sum_{\tau \in \mathcal{T}} e^{-\beta\,(s-\tau)}$
	 \STATE $u \sim \mbox{uniform}(0,1)$
	 \STATE $w \gets -\log(u)/\bar{\lambda}$
	 \STATE $s \gets s+w$
	 \STATE $D \sim \mbox{uniform}(0,1)$
	 \IF{$D\,\bar{\lambda} \leq \lambda(s) = \lambda_o + \alpha\,\sum_{\tau \in \mathcal{T}}e^{-\beta\,(s-\tau)}$} 
	 \STATE $n \gets n+1,\, t_n \gets s,\, \mathcal{T} = \mathcal{T} \bigcup \{t_n\}$
	 \ENDIF
   \ENDWHILE 
	 \IF{$t_n \leq T$} 
	 \STATE {\bf return} $\{\tau_o+t_k\}_{k=1,2,.\,.\,.,n}$  
	 \ELSE \STATE {\bf return} $\{\tau_o+t_k\}_{k=1,2,.\,.\,.,n-1}$
	 \ENDIF
\end{algorithmic}
\end{algorithm}

\newenvironment{myequation2}{%
\addtocounter{equation}{-7}
\refstepcounter{defcounter}
\renewcommand\theequation{C\thedefcounter}
\begin{equation}}
{\end{equation}}

\subsection*{Appendix C: Pseudo-codes for the Learning and Inference Algorithms}
In this Subsection, we provide the detailed implementation of the learning and inference algorithms presented in Sections \ref{rscore} and \ref{Offlearn}. We start first by presenting the real-time risk scoring algorithm in the following Subsection. 
\subsubsection*{The Real-time Risk Scoring Algorithm}
We assume that we know the model parameters $\hat{\Omega}$ (in practice, this will be learned from the offline dataset $\mathcal{D}$). As explained earlier in Sections \ref{rscore}, the patient risk score at time $t$ is denoted by $R(t)$, and is defined as follows
\begin{myequation2}
R(t) = \mathbb{P}(X(\infty) = N\,|\,\{y_m,t_m\}, t_m \leq t, \Omega).
\label{eqC1}
\end{myequation2}
In other words, the risk score $R(t)$ is the probability of the patient's eventual absorption in the deteriorating state $N$ given the observable physiological data up to time $t$. The expression in (\ref{eqC1}) can be decomposed using Bayes' rule as follows  
\begin{myequation2}
R(t) = \sum_{i \in \mathcal{X}} \mathbb{P}(X(t) = i|\{y_m,t_m\})\mathbb{P}(X(\infty) = N|X(t)=i).
\label{eqC2}
\end{myequation2}
Given the model parameters $\Omega$, one can easily compute the transition probabilities $\mathbb{P}(X(t+\Delta t) = j|X(t)=i)$ for any $\Delta t$, $i$ and $j$ using the parameter the semi-Markov transition kernel and the sojourn time distributions. (For a semi-Markov chain, the transition probabilities are the solutions of a Volterra integral equation, which parallels the Chapman-Kolomogrov equations in ordinary Markov chains.) Now we focus on the term $\mathbb{P}(X(t) = i|\{y_m,t_m\})$: the posterior distribution of the current state given the observations up to the current time. We compute this term using {\it forward-filtering}. Define the {\it forward message} as    
\begin{myequation2}
\alpha(t_m, i) = \mathbb{P}(X(t) = i, \{y_m,t_m\}).
\label{eqC3}
\end{myequation2}
The forward messages in (\ref{eqC3}) are computed efficiently using dynamic programming as in \cite{yu2010hidden}. The risk score is thus computed as follows
\begin{myequation2}
R(t) = \sum_{i \in \mathcal{X}} \frac{\alpha(t_m, i)}{\sum_{j}\alpha(t_m, j)}\,\cdot\,\mathbb{P}(X(\infty) = N|X(t)=i).
\label{eqC4}
\end{myequation2}

\subsubsection*{The Offline Learning Algorithm}
Now we describe the offline learning algorithm by going through steps 1, 2 and 3 discussed in Section \ref{Offlearn} in detail.\\
\\ 
{\bf Step 1:}\\
We estimate the jump times $\{\tau_1,.\,.\,.,\tau_{K}\}$ for an episode $\mathcal{E}$ by using the {\it E-divisive} nonparametric mutli-variate change-point detection algorithm \cite{matteson2014nonparametric}. We jointly detect the changes in the distributions of the observable variables $\{y^d_m\}^{M_d}_{m=1}$ and the observation process $\{t^d_m\}^{M_d}_{m=1}$ by creating an {\it augmented} vector of observables that comprises both the physiological observations and a ``differential" observation process as follows: 
\[\left(\begin{pmatrix}
  y_1(1)  \\
	y_1(2)  \\
  \vdots   \\
  y_1(Q)  \\ 
	\Delta t_1  \\ 
 \end{pmatrix},
.\,.\,.,
\begin{pmatrix}
  y_m(1)  \\
	y_m(2)  \\
  \vdots   \\
  y_m(Q)  \\ 
	\Delta t_m  \\ 
 \end{pmatrix},
.\,.\,.,
\begin{pmatrix}
  y_M(1)  \\
	y_M(2)  \\
  \vdots   \\
  y_M(Q)  \\ 
	\Delta t_M \\ 
 \end{pmatrix}\right)
\]
where $\Delta t_m = t_m - t_{m-1}$, with $\Delta t_1 = 0$. We run the E-divisive algorithm with a significance level of 0.05 for the permutation test and a parameter setting $\alpha = 1$ for the moment distance. The output of this step is a set of jump time estimates $\{\hat{\tau}_1,.\,.\,.,\hat{\tau}_K\}$.\\ 
\\
{\bf Step 2:}\\
Now that we have estimates for the onsets of the absorbing state $\hat{\tau}_K$, obtained from step 1, then we can estimate the response times as $\hat{S}_{K} = T_c-\hat{\tau}_{K}$. Let us assume that we have a set of $D$ episodes absorbed in state 1 for whom the estimated response times are $\{\hat{S}^d_{K}\}_{d=1}^D$ with labels $\{l^d = 0\}_{d=1}^D$, and let the Gamma distribution parameters be $\gamma_1 = \{k_1, \beta_1\}.$ Then the MLE of $\gamma_1$ and $\beta_1$ as follows   
\begin{align}
v &= \log\left(\frac{1}{D}\sum_{i=1}^{D}\hat{S}^i_{K}\right)-\frac{1}{D}\sum_{i=1}^{D}\log(\hat{S}^i_{K}) \nonumber \\
\hat{k}_1 &= \frac{3-v+\sqrt{(v-3)^2+24\,v}}{12\,v} \nonumber\\
\hat{\beta}_1 &= \frac{1}{\hat{k}_1\,D}\sum_{i=1}^{D}\hat{S}^i_{K}, \nonumber
\label{eqC5}
\end{align}
and the same estimates are found for $\gamma_N = \{k_N, \beta_N\}.$ 

Now let $\{y_m, t_m\}$ be the sampling times and observed variables within the response time interval $[\hat{S}_{K}, T_c]$. For an episode with absorbed in state $i \in \{1,N\}$, we estimate the multi-task GP parameter $\Theta_i$ using the gradient method as in \cite{bonilla2007multi}. The Hawkes process parameters $\Lambda_i$ are estimated by maximizing the recursive likelihood formula in \cite{ogata1981lewis}:
\begin{align}
\log(\mathbb{P}(\{t_m\}^{M}_{m=1}|\Lambda_i)) &= -\lambda^o_i\,t_m + H(m,\Lambda_i), \nonumber
\end{align}
\[H(m,\Lambda_i) =\] 
\begin{align}
\sum_{m=1}^{M}\left(\frac{\alpha_i}{\beta_i}(e^{-\beta_i(t_n-t_m)}-1)+\sum_{m=1}^{M}\log(\lambda^o_i + \alpha^i\,A(m))\right),\nonumber
\end{align}
where $A(m) = \sum_{t_j < t_m} e^{-\beta_i(t_m-t_j)}$. The maximization is conducted using the Nelder-Mead simplex method \cite{nelder1965simplex}, implementation is done through the \texttt{nlm} function in \texttt{R}. \\
\\
{\bf Step 3:}\\
After segmenting the episodes using the estimated jump times $\{\tau_1,.\,.\,.,\tau_{K}\}$ in step 1, we view the segment episode as a discrete-time HMM with no self-transition and with transition probabilities $[p_{ij}]_{ij}$ and initial state probabilities $[\pi_i]_i$. The observations associated with every segment is a set of observation times $\{t_m\}$, the observations $\{y_m\}$ and the segment duration $\hat{\tau}_{n}-\hat{\tau}_{n-1}$. The transition states' parameters are then estimated straightforwardly using the Baum-Welch algorithm. 

\subsection*{Appendix D: Data Description and Implementation of Baseline Algorithms}
\subsubsection*{Data Description}
We conducted our experiments on a very heterogeneous cohort of episodes for patients hospitalized in a major medical center over the last 3 years. The cohort involved all the hospital's major units, namely, the cardiac observation floor, the cardiothoracic floor, the hematology and stem cell transplant floor, the liver transplant service and the critical care pediatrics unit. The cohort involved patients who were undergoing narcotic drugs or chemotherapy, and hence are very are vulnerable to adverse outcomes that require an impending ICU transfer. The patients' cohort displays vast heterogeneity in terms of a wide variety of diagnoses and ICD-9 codes including leukemia, hypertension, septicemia, sepsis, pneumonia, and renal failure. The distribution of the most frequent ICD-9 codes in the patient cohort (together with the corresponding diagnoses) is provided in Table \ref{TabApp1}. 

\begin{table}
    \centering
\begin{tabular}{|c|c|c|}
\toprule[1.25pt]
{\bf ICD-9 codes} & {\bf Diagnosis} & {\bf $\%$ Freq.} \\ 
\hline
\hline
{\bf (786.05)} & Shortness of Breath & 7$\%$\\ 
{\bf (401.9)} & Hypertension & 6$\%$\\ 
{\bf (38.9)} & Septicemia & 5$\%$\\ 
{\bf (995.91)} & Sepsis & 5$\%$\\ 
{\bf (780.6)} & Fever & 5$\%$\\ 
{\bf (486)} & Pneumonia & 5$\%$\\   
{\bf (584.9)} & Renal failure & 5$\%$\\ 
{\bf (599)} & Urethra and urinary attack & 5$\%$\\ 
{\bf (780.97)} & Altered mental status & 4$\%$\\ 
{\bf (285.9)} & Anemia & 4$\%$\\ 
{\bf (786.5)} & Chest pain & 4$\%$\\
{\bf (585)} & Chronic renal failure & 4$\%$\\ 
{\bf (780.79)} & Malaise and fatigue & 3$\%$\\
{\bf (578)} & Gastrointestinal hemorrhage & 3$\%$\\  
{\bf (428)} & Heart failure & 3$\%$\\ 
{\bf (427.31)} & Atrial fibrillation & 3$\%$\\ 
{\bf (787.01)} & Nausea &  3$\%$\\
{\bf ---} & Other & 22.5$\%$\\
\hline
\bottomrule[1.25pt]
\end{tabular}
\label{TabApp1}
\end{table}
Every patient's episode in the cohort is associated with a set of 21 (temporal) physiological streams comprising a set of vital signs and lab tests that are listed as follows.\\
\\ 
{\bf Vital signs}
\begin{itemize}
\item Diastolic blood pressure
\item Systolic blood pressure
\item Glasgow coma scale score
\item Heart rate
\item Eye opening 
\item Respiratory rate
\item Temperature 
\item $O_2$ saturation and device assistance
\item Best motor and verbal responses.
\end{itemize}
\newpage
{\bf Lab tests}
\begin{itemize}
\item Chloride
\item Glucose
\item Urea nitrogen
\item White blood cell count
\item Creatinine
\item Hemoglobin
\item Platelet count
\item Potassium
\item Sodium 
\item $CO_2$
\end{itemize}
In all the experiments, we split the patient cohort into a training set, covering episodes for a 2.5 year period, and a testing set, covering episodes recorded over a 6 month period. The training set comprises 5,000 episodes, whereas the testing set comprises 1,094 episodes. All the patient episodes in the cohort were informatively censored. That is, for every patient in the cohort, we know the following information: the censoring time $T_c$, i.e. the length of stay of each patient in the ward. The average hospitalization time (or censoring time) in the ward was 150 hours and 24 minutes. The patient episodes' censoring times ranged from 4 hours to more than 2,500 hours. We also know the absorbing clinical state ($l$) for every patient.

\subsubsection*{Implementation of the Baselines}
In this Section, we present the details of our implementation for all the baseline algorithms involved in the comparisons in Section \ref{Expr}. We first start by explaining how the state-of-the-art clinical risk scores were implemented.\\ 
\\
\textit{\textbf{Implementation of the Clinical Risk Scores}}
\begin{enumerate}[(1)]
\item {\bf Rothman Index}: we implement the stepwise logistic regression scheme adopted by the Rothman index as described in \cite{rothman2013development}. The Rothman risk score was computed as in equation (1) and (2) in \cite{rothman2013development}, i.e.
\[\mbox{Rothman Index} = \left[\mbox{RI}_{\mbox{no lab}}\left(\frac{\mbox{Time since lab}}{48}\right)\right] +\]
\[\mbox{Smoothing function}\, \left[\mbox{RI}_{\mbox{lab}}\left(1-\frac{\mbox{Time since lab}}{48}\right)\right],\]
where the ``RI" for every lab test (or vital sign) are obtained from Figure A1 in \cite{rothman2013development} for the entire 21 vital signs and lab tests available in the EHR, and the smoothing function is obtained from Appendix B in \cite{rothman2013development}. The ``Time since lab" variable is the last time (in hours) in which the corresponding lab test (or vital sign) was gathered, and its maximum value is 48 hours. At the time of conducting these experiments, the Rothman index was deployed in the medical center from which we obtained the data.
 
\item {\bf Modified Early Warning System (MEWS)}: we implemented MEWS as specified in \cite{morgan1997early}. The MEWS score typically ranges from 0 to 3 and is computed over time based on the instantaneous values of the following cardinal vital signs: systolic blood pressure, respiratory rate, SaO2, temperature, and heart rate. Table \ref{Tab2ICU} provides the MEWS risk scoring function in terms of those vital signs.
\begin{table*}[t]
\centering
\caption{\small Computation of the MEWS score}
\begin{tabular}{|c|c|c|c|c|c|c|c|}
\toprule[1.25pt]
{\bf Score} & {\bf 3}	& {\bf 2}	& {\bf 1}	& {\bf 0}	& {\bf 1}	& {\bf 2}	& {\bf 3} \\ 
\hline
\hline
Heart rate (bpm) & $>129$ & $110-129$ & $100-109$ & $50-99$ & $40-49$ & $30-39$ & $<30$\\ \hline
Temperature (C)	 & $---$ & $>38.9$ & $38-38.9$ & $36-37.9$ & $35-35.9$ & $34-34.9$ & $<34$ \\ \hline
Systolic BP (mmHg) & $---$ & $>199$ & $---$	& $100-199$ &	$80-99$ & $70-79$	& $<70$ \\ \hline
SpO2 ($\%$)	& $<85$ &	$85-89$	& $90-92$	& $>92$ & $---$ & $---$ & $---$	\\ \hline
Respiratory rate (breaths/min)	& $>35$ & $31-35$ & $21-30$ & $9-20$ & $---$ & $---$ & $<7$ \\
\hline
\bottomrule[1.25pt]
\end{tabular}
\label{Tab2ICU}
\end{table*}
\item {\bf Sequential Organ Failure Assessment (SOFA)}: a risk score (ranging from 1 to 4) that is used to determine the extent of a hospitalized patient's respiratory, cardiovascular, hepatic, coagulation, renal and neurological organ function in the ICU.
\item {\bf Acute Physiology and Chronic Health Evaluation (APACHE-II)}: a risk scoring system (an integer score from 0 to 71) for predicting mortality of patients in the ICU. The score is based on 12 physiological measurements, including creatinine, white blood cell count, and Glasgow coma scale.
\end{enumerate}
We note that while the SOFA and APACHE II scores were originally constructed for deployed for patients in the ICU, both scores have been recently shown to provide a prognostic utility for predicting clinical deterioration for patients in regular wards \cite{yu2014comparison}, and hence we consider both scores in our comparisons. Our implementation for APACHE-II and SOFA followed that in \cite{yu2014comparison}.\\ 
\\
\textit{\textbf{Implementation of the machine learning baselines}} \\
\\
We have not used any of the patients' static features (e.g. age, gender, diagnoses, etc) in the baselines to ensure a fair comparison with the medical risk scores, which ignore those features. For all the baseline (including our algorithm), the relevant physiological time series for every baseline were selected through the {\it correlated feature selection} method \cite{yu2003feature}. The hyper-parameters of all the baselines, including the size of the sliding window for the supervised learning methods, were optimized via cross-validation. In order to handle the irregularly sampled data, we discretized the time horizon into 1-hour steps and fed the baselines with spline-interpolated, discrete-time episodes. 

In the following, we provide the implementation details for the different machine learning baselines. 
\begin{enumerate}[(1)]
\item {\bf Hidden Markov Model with Gaussian Emissions}: informative censoring information was incorporated by including two absorbing HMM states for clinical stability ($l=0$) and deterioration ($l=1$). We used the Baum-Welch algorithm for learning the HMM, and informed the forward-backward algorithm with the labeled states $l$ at the end of every episode. We initialized the Baum-Welch algorithm with seed parameter estimates that are obtained using a $K$-means clustering of the patients' episodes followed by change-point detection (using E-divisive) and then MLE for a labeled HMM. The complete data log likelihood after 200 EM iterations was -5.35$\times$ 10$^6$. Using the Bayesian information criterion, we selected an HMM with 5 states (2 of which are absorbing). In the testing phase, a patient's risk score at every point of time is computed by first applying forward filtering to obtain the posterior probability of the patient's states, and then averaging over the distribution of the absorbing states. 

\item {\bf Sliding Window Methods}: in order to ensure that the censoring information is utilized by all the sliding-window predictors, we trained every predictor by constructing a training dataset that comprises the physiological data gathered within a temporal window before the censoring event (ICU admission or patient discharge), and using the censoring information (i.e. the variable $l$) as the labels. The size of this window is a hyper-parameter that is tuned separately for every predictor. For the testing data, the predictors are applied sequentially to a sliding window of every patient's episode, and the predictor's output is considered as the patient's real-time risk score. This differs from the static simulation setting in \cite{ghassemi2015multivariate} were predictions are issued in a one-shot fashion using only the data obtained within 24 hours after a patient's admission. We used the built-in MATLAB functions for training the logistic regression and the random forest predictors. For SW-GP, we used multi-task Gaussian process regression using squared exponential kernel and using the free-form parametrization in \cite{bonilla2007multi}, and used the gradient method to learn the parameters of two Gaussian process models: one for patients with $l=0$, and one for patients with $l=1$. The risk score for a patient's risk score is computed as the test statistic of a sequential hypothesis test that is based on the two learned Gaussian process models as in \cite{ForecastICU}. Hence, the SW-GP baseline is a combination of the methods in \cite{ghassemi2015multivariate} and \cite{ForecastICU} that is capable of handling irregularly sampled data.   
\end{enumerate}

\end{document}